# Rule reasoning for legal norm validation of FSTP facts

Naouel Karam[1], Shashishekar Ramakrishna[1, 2] and Adrian Paschke[1]
[1]AG Corporate Semantic Web, Department of Computer Science Freie Universitat Berlin, Germany
[2] TelesPRI GmbH Berlin, Germany
{naouel.karam@fu-berlin.de}

Extended Abstract

Non-obviousness or inventive step is a general requirement for patentability in most patent law systems. An invention should be at an adequate distance beyond its prior art in order to be patented. Fulfilling a minimum measurement limit would enable a patent applicant to have its invention patented. Based on this fact, we proposed a method for nonobviousness analysis of a patent over its prior arts, based on highest court's precedents, called the FSTP[1] Test [1] [2].

The FSTP Test works on formal descriptions of the technique teaching. To derive this formal representation of the technique teaching we need to identify different kinds of information explicitly or implicitly disclosed in the patent description and its Reference Set (RS). This formal representation identifies three main discretization levels: elements, attributes and concepts. *Elements* represent the important characterizing aspects of an invention, a*ttributes* represent the properties of the elements and *concepts* represent the elementary properties referred by the attributes.

The proposed method elaborates on several compliance tests defined by different National Patent Systems (NPS) on patent nonobviousness analysis and subject matter eligibility. From the scope of this paper, we consider tests (U.S.C 35 §101, §102, §103, §112 …) [3] defined by the U.S. Supreme Court and Court of Appeals for the Federal Circuit (CAFC), and the order in which such test are to be applied as depicted by U.S CAFC, in its recent decisions of, RETRACTABLE TECH [4] and PHILIPS [5]. Furthermore, U.S CAFC also highlighted the imminent need for discretization levels before applying such tests.

For transforming facts, indentified on different discretization levels into the quantified semantic facts, the FSTP test provides a platform for applying series of compliance tests from the Title 35 of United States Code for patent law (35 U.S.C) to support its existing semantic and pragmatic layers. On the first stage, norms pertaining to U.S.C 35 §112 are applied to determine the claimed inventions inventive concepts and thereby, its semantics over its prior art. Subsequently, U.S.C 35 §102/103 tests for determining the claimed inventions patentability semantics as indicated by the facts are applied. Finally facts that qualify the first and the second stages of tests are subject to U.S.C 35 §101 compliance test for its patent-eligible pragmatics. To sum up, all semantic/creative facts need to be validated in a given order and in accordance to the respective NPS under consideration for justifying the pragmatic/ innovative height of the patent's invention over its prior art.

[1] This work has been supported by the Fact Screening and Transforming Processor (FSTP) project funded by the Teles Pri GmbH: www.fstp-expert-system.com

In order to perform such compliance tests, legal norms need to be transformed from their natural language format to a formal representation format. We propose to use LegalRuleML [6], an XML standard for legal knowledge representation based on RuleML [7] and which supports the modelling of norms. We use it for representing patent norms and their case law decisions as legal rules. It resolves complex legal questions and automates the analysis of a large number of patent norms with respect to their logical coherence in a given NPS. It can also be used to point out logical inconsistencies in current case law decisions and can be used to evaluate compliance of semantic facts with case law and positive law.

For legal reasoning, we propose to convert LegalRuleML into RuleML, which then can be evaluated using a rule reasoner as shown in Fig 1.

The formal FSTP facts identified in the first layer need to be annotated in order to allow the reasoner to evaluate the corresponding rules. We use both linguistic and ontological information as a base for annotations in connection with the rules semantics.

Such system supports life cycle management of the knowledge. It captures the changes over time of the rules when the legal binding text changes. Updates in the NPSs by new decisions will lead to corresponding isomorphic updates in the NPSs knowledge bases.

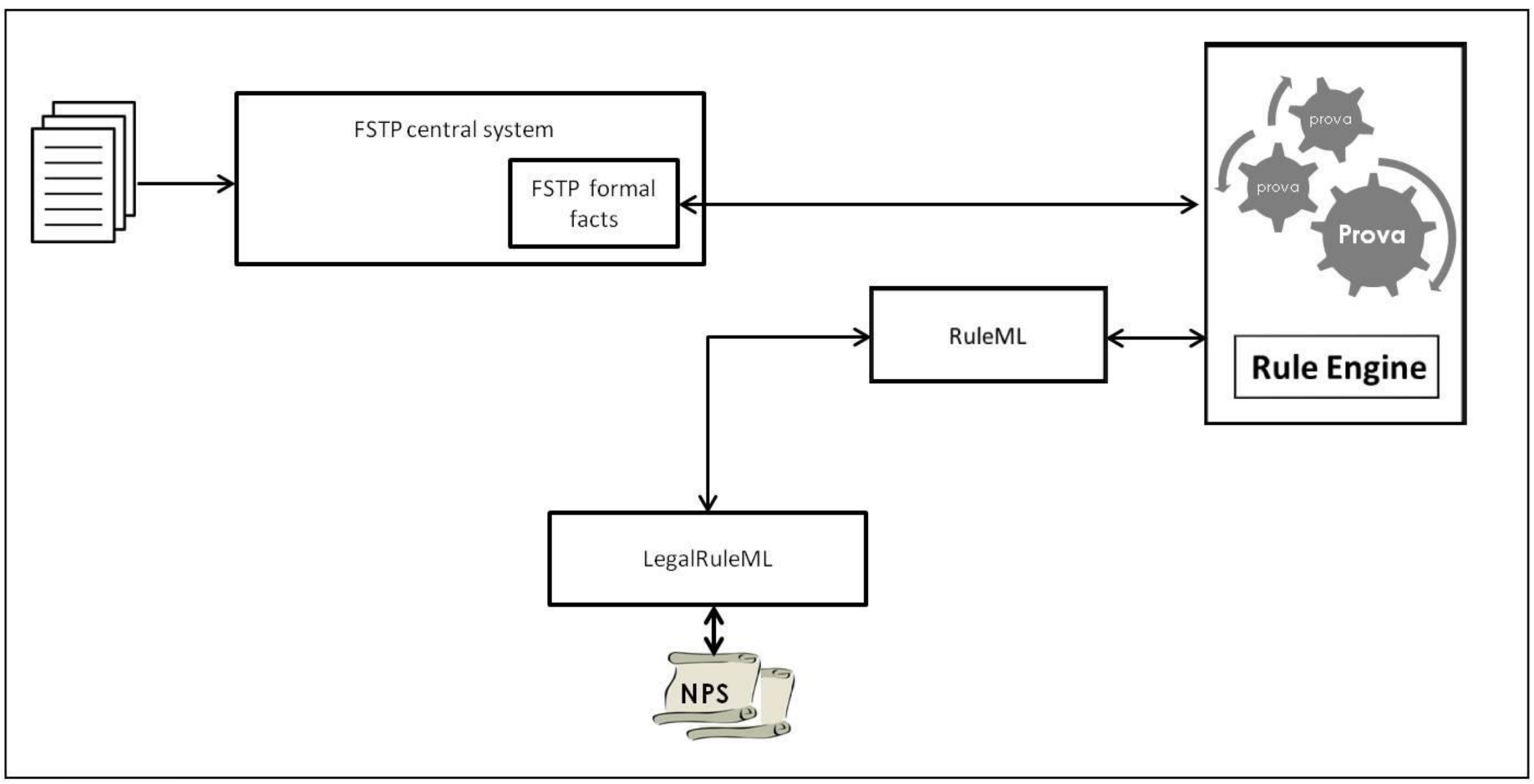

**Figure 1 : Legal norm validation of FSTP facts using rule reasoning**

We are implementing a prototype in conjunction with the FSTP expert system [1] for rule based patent norm reasoning. In particular, we are addressing specific sections from national patent system connected to nonobviousness analysis. For rule reasoning we use Prova [8], a java based open source rule language for reactive agents and event processing.